\documentclass[letterpaper, 10 pt, conference]{ieeeconf}  

\IEEEoverridecommandlockouts                              

\usepackage{graphicx}
\usepackage{amsmath}
\usepackage{amssymb}
\usepackage{mathtools}
\usepackage{array}
\usepackage{booktabs}
\usepackage{multirow}
\usepackage{tabularx}
\usepackage{url}
\usepackage[capitalise]{cleveref}

\crefname{equation}{}{}
\Crefname{equation}{}{}
\DeclareMathOperator{\diag}{diag}
\DeclareMathOperator*{\argmin}{arg\,min}

\newcolumntype{Y}{>{\centering\arraybackslash}m{\dimexpr\linewidth/6\relax}}

\title{\LARGE \bf
ORN-CBF: Learning Observation-conditioned Residual Neural Control Barrier Functions via Hypernetworks
}

\author{Bojan Derajić$^{1, 2}$, Sebastian Bernhard$^{1}$ and Wolfgang Hönig$^{2}$
\thanks{$^{1}$AUMOVIO, Germany }%
\thanks{$^{2}$Technical University of Berlin, Germany}%
\thanks{Contact e-mail: {\tt\small bojan.derajic@aumovio.com}}%
\thanks{This work is funded by the German Federal Ministry for Economic Affairs and Climate Action within the project \textit{nxtAIM}.}%
}

\begin{document}

\maketitle
\thispagestyle{empty}
\pagestyle{empty}

\begin{abstract}

Control barrier functions (CBFs) have been demonstrated as an effective method for safety-critical control of autonomous systems. Although CBFs are simple to deploy, their design remains challenging, motivating the development of learning-based approaches. Yet, issues such as suboptimal safe sets, applicability in partially observable environments, and lack of rigorous safety guarantees persist. In this work, we propose observation-conditioned neural CBFs based on Hamilton-Jacobi (HJ) reachability analysis, which approximately recover the maximal safe sets. We exploit certain mathematical properties of the HJ value function, ensuring that the predicted safe set never intersects with the observed failure set. Moreover, we leverage a hypernetwork-based architecture that is particularly suitable for the design of observation-conditioned safety filters. The proposed method is examined both in simulation and hardware experiments for a ground robot and a quadcopter. The results show improved success rates and generalization to out-of-domain environments compared to the baselines.

\end{abstract}

\section{INTRODUCTION}
\label{sec:introduction}

One of the most prominent and well-established approaches to safety-critical control is the so-called \textit{safety filtering}. This approach functions by monitoring the nominal control (obtained from, e.g., a performance-oriented controller or a human operator) and modifying it when necessary to maintain safe operation of the system. The nominal control is usually modified by a switching mechanism or through an optimization procedure \cite{hsu_safety_2024}. In this paper, we focus on the second approach, which specifically utilizes control barrier functions (CBFs).

CBFs, inspired by control Lyapunov functions, are scalar functions defined over the system's state space that characterize control invariant sets based on Nagumo's Theorem on set invariance \cite{nagumo_uber_1942, ames_control_2019}. In practice, the key advantage of using CBFs is emphasized for control-affine systems, where the safety filtering is reduced to solving a quadratic program (QP) and can be performed with high frequency \cite{ames_control_2014}. Although CBFs provide an elegant way to enforce safety, the main difficulty lies in designing a proper CBF. This problem is especially complex for nonlinear systems with state and input constraints, and most existing approaches make assumptions or simplifications to obtain a solution \cite{hsu_safety_2024}. Also, an additional layer of complexity is present for systems such as mobile robots that operate in unknown environments based on local observations. For such systems, the CBF must be generated in real time using the available observations, meaning that the existing methods for offline CBF design are not applicable \cite{harms_neural_2024}.

\begin{figure}[!htb]
    \centering
    \includegraphics[width=\linewidth]{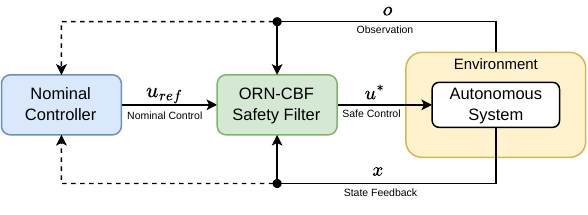}
    \caption{Safety filtering for autonomous systems operating based on environment observations with the proposed ORN-CBF method.}
    \label{fig:safety_filtering}
\end{figure}

In this paper, we propose a learning-based approach for designing observation-conditioned CBFs that overcome the limitations of existing methods by leveraging a few key features. First, we use an efficient hypernetwork-based architecture in which the hypernetwork parametrizes the main network given available observations (2D occupancy grid in our experiments), while the main network approximates the target function over the system's state space. This architecture improves efficiency by inferring the hypernetwork (complex model) only once when a new observation becomes available, while the main network (simple model) is queried frequently for its value and gradient. Second, we use Hamilton-Jacobi (HJ) reachability analysis as the source of supervision during training, resulting in CBFs that approximately recover the maximal control invariant set \cite{choi_robust_2021, tonkens_refining_2022, so_how_2024}. Finally, instead of directly approximating the HJ value function, the main network approximates only the residual component with respect to the signed distance function (SDF). Since the residual function is always nonnegative \cite{derajic_residual_2025}, by applying a nonnegative activation function at the output of the main network, the safe set associated with the predicted CBF never intersects with the observed failure set. The summary of the contributions is:
\begin{itemize}
    \item A novel observation-conditioned neural CBF for safe navigation of mobile robots in unknown environments with arbitrary distribution and shapes of obstacles. The proposed method approximately recovers the optimal safe set and it guarantees, by design, that the predicted safe set does not intersect with the observed failure set.
    \item Extensive evaluation of the proposed ORN-CBF (\cref{fig:safety_filtering}) method for a ground robot and a quadcopter, both in 3D simulations and hardware experiments, demonstrating its performance and application flexibility.
\end{itemize}

\section{RELATED WORK}
\label{sec:related_work}

There is a significant amount of work on CBFs published in recent years as they have been proven to be a powerful, yet elegant way to enforce safety \cite{ames_control_2019}.  Numerous researchers contributed through their work on different variants of CBFs, including exponential, discrete-time, robust, time-varying, and adaptive CBFs \cite{garg_advances_2024}. Nevertheless, designing a proper CBF for a general nonlinear system with state and control constraints remains the major challenge in the field \cite{hsu_safety_2024}.

The particularly relevant stream of research leverages machine learning (ML) to design CBFs in a more flexible and scalable way. For example, supervised ML is applied in \cite{srinivasan_synthesis_2020} to construct CBFs from LiDAR rays, learning CBFs from expert trajectories is proposed in \cite{robey_learning_2020}, while the neural CBFs for systems with limited control are introduced in \cite{liu_safe_2022}. Also, an uncertainty-aware CBF method based on Gaussian processes is developed in \cite{li_robust_2023}. However, these approaches often fail to recover the maximal safe set, which can be obtained via HJ reachability analysis \cite{so_how_2024}.  The connection between CBFs and HJ reachability is established in \cite{choi_robust_2021}, where the novel control-barrier value function (CBVF) is introduced. This approach is further extended in \cite{tonkens_refining_2022} to iteratively refine candidate CBFs, hand-crafted or learned, to enlarge the initial safe set. Similarly, authors in \cite{tonkens_patching_2024} use HJ reachability to locally modify potentially unsafe regions associated with a neural CBF, improving the overall scalability. Still, these methods have been demonstrated only for systems operating in known environments, which is the key difference to our method. 

Regarding the design of neural CBFs in unknown environments, the method in \cite{long_learning_2021} learns a neural SDF approximation for the given observation and uses it as a CBF. The main drawback is that the training is performed online, limiting its application to systems with slower dynamics. Differently, the authors in \cite{dawson_learning_2022} use a discrete-time CBF, which requires predicting the sensor observation one time step into the future. This approach is feasible for simple sensors such as 2D laser scan, but might be impractical for more complex modalities. On the other hand, authors in \cite{harms_neural_2024} train a neural model to approximate the value function associated with the state-dependent Riccati equation based on a 2D laser scan and use it as CBF.

The common downside of the observation-based CBFs mentioned above is that they fail to recover the optimal safe set and lack any safety guarantees. Similar to \cite{derajic_learning_2025} and \cite{derajic_residual_2025} that employ a neural approximation of the HJ value function for model predictive control (MPC) design, we use such an approximation as a CBF. We also exploit certain mathematical properties of the HJ value function, allowing us to guarantee that the resulting safe set never includes the failure set. Moreover, we leverage a hypernetwork-based architecture, making our method particularly efficient for safety filtering based on observations.


\section{PRELIMINARIES}
\label{sec:preliminaries}

\subsection{System Dynamics}
\label{subsec:system_dynamics}

In this paper, we consider continuous-time systems with control-affine dynamics:
\begin{equation} \label{eq:sys_dynamics}
    \dot{x}= f_c(x, u) = f(x) + g(x) u,
\end{equation}
where ${x \in \mathcal{X} \subseteq \mathbb{R}^n}$ is the state vector and ${u \in \mathcal{U} \subseteq \mathbb{R}^m}$ is the control input. We assume that $f_c$ satisfies the standard requirements on the solution existence and uniqueness, and that full state feedback is available.

\subsection{Control Barrier Functions}
\label{subsec:cbf}

A continuously differentiable function $h: \mathcal{X} \rightarrow  \mathbb{R}$ is a \textit{control barrier function (CBF)} if there exists an (extended) class $\kappa_{\infty}$ function $\alpha$ such that for the system \cref{eq:sys_dynamics}
\begin{equation} \label{eq:cbf}
    \sup_{u \in \mathcal{U}} \nabla h(x)^\top f_c(x, u) \geq -\alpha \left( h(x) \right), \, \forall x \in \mathcal{X}.
\end{equation}
If we define set $\mathcal{S}$ as
\begin{equation} \label{eq:safe_set}
    \mathcal{S} = \{  x \in \mathcal{X} : h(x) \geq 0 \},
\end{equation}
then $\mathcal{S}$ is forward invariant and we call it the \textit{safe set} \cite{ames_control_2019}. 

Taking into account that the system is control-affine, based on the Lie derivatives $L_{f}h(x)$ and $L_{g}h(x)$ with respect to $f(x)$ and $g(x)$, respectively, the inequality \cref{eq:cbf} can be rewritten as
\begin{equation} \label{eq:cbf_control_affine}
    \sup_{u \in \mathcal{U}} \left[ L_{f}h(x) + L_{g}h(x) u \right] \geq -\alpha \left( h(x) \right), \, \forall x \in \mathcal{X},
\end{equation}
which is linear in $u$ for a given $x$. Moreover, if the set of admissible controls $\mathcal{U}$ can be represented by a set of linear constraints, the CBF condition \cref{eq:cbf_control_affine} can be used to design a safety filter in the form of a quadratic program (CBF-QP) \cite{hsu_safety_2024, ames_control_2014}. The CBF-QP safety filter provides safe control $u^*$ by solving the following optimization problem for the nominal control $u_{ref}$ and state $x$:
\begin{equation} \label{eq:cbf_qp}
\begin{aligned}    
    u^* = \argmin_{u \in \mathcal{U}} \,\, &\frac{1}{2} \| u_{ref} - u \|^2 \\
    s.t. \quad &L_{f}h(x) + L_{g}h(x) u + \alpha \left( h(x) \right)  \geq 0. 
\end{aligned}
\end{equation}
Thanks to modern QP solvers, CBF-QP can run at relatively high frequencies even with scarce computational resources, and we will design such a safety filter in this paper.

\subsection{Hamilton-Jacobi Reachability Analysis}
\label{subsec:hjr}

HJ reachability analysis is a framework for verification and analysis of dynamical systems grounded in the optimal control theory \cite{bansal_hamilton-jacobi_2017}. In this paper, we consider undisturbed systems described by \cref{eq:sys_dynamics}. If the system starts at state $x(t)$, then $\xi^{u}_{x, t}(\tau)$ is the system's state at time $\tau$ after applying $u(\cdot)$ over time horizon $[t, \tau]$. The set of states that should be ultimately avoided is called a failure set $\mathcal{F}$. Also, we define a backward reachable tube (BRT) $\mathcal{B}(t)$, which is a set of initial states from which the system will reach $\mathcal{F}$ within the time horizon $[t, T]$ for any control $u(\cdot)$, i.e. 
\begin{equation}
    \mathcal{B}(t) = \{ x: \forall u(\cdot), \, \exists \tau \in [t, T], \, \xi^{u}_{x, t}(\tau)  \in \mathcal{F} \}.
\end{equation}

To compute the BRT for a given failure set, we define $\mathcal{F}$ as the zero-sublevel set of a failure function $F(x)$, i.e. ${\mathcal{F} = \{x:F(x) \leq 0 \}}$. The computation of the BRT can be formulated as an optimization problem where the objective is the minimal distance to $\mathcal{F}$ over the time horizon, i.e.
\begin{equation}
    J(x, t, u(\cdot)) = \min_{\tau \in [t, T]} F(\xi^{u}_{x, t}(\tau)).
\end{equation}
Since $\mathcal{F}$ represents an unsafe region, the goal is to find the optimal control that will maximize this distance and therefore we introduce the following value function:
\begin{equation} \label{eq:value_func}
    V(x, t) = \sup_{u(\cdot) \in \mathcal{U}} \{ J(x, t, u(\cdot)) \}.
\end{equation}
This value function can be computed using the dynamic programming principle to solve the following Hamilton-Jacobi-Bellman Variational Inequality (HJB-VI) \cite{bansal_hamilton-jacobi_2017}:
\begin{equation} \label{eq:hjb_vi}
    \begin{aligned}
    \min \left\{ \frac{\partial}{\partial t}V(x, t) + H(x, t), \, F(x) - V(x, t) \right\} &= 0, \\
    V(x, T) &= F(x).
    \end{aligned}
\end{equation}
In the formulation above, Hamiltonian $H(x, t)$ is defined as
\begin{equation}
    H(x, t) = \max_{u \in \mathcal{U}} \nabla V(x, t)^\top f_c(x, u).
\end{equation}
Once the value function \cref{eq:value_func} is obtained, the corresponding BRT can be described as its zero-sublevel set, i.e.
\begin{equation}
    \mathcal{B}(t) = \left\{ x: V(x, t) \leq 0 \right\},
\end{equation}
and the maximal safe set is its complement: $\mathcal{S}(t) = \mathcal{B}(t)^c$.

In this paper, the failure function is the space occupied by obstacles and defined as an SDF $d(x)$, i.e., $F(x) \coloneq d(x)$. Also, by assuming a terminal time $T=0$ and a static environment, we propagate the value function backward in time until it converges to its steady-state value $V(x) = \lim_{t\rightarrow-\infty}V(x, t)$ and then use $V(x)$ as the target CBF during training\footnote{Although HJ value function is a viscosity solution to the HJB-VI and theoretically might not be differentiable everywhere, the gradients can always be computed via numeric or automatic differentiation in practice.}.

\section{METHODOLOGY}
\label{sec:methodology}

\subsection{Observation-conditioned CBFs}
\label{subsec:obs_cond_cbf}

When dealing with autonomous systems operating based on the exteroceptive observations of the environment, the CBF is generally a function of the system's state $x$ and observation $o$, i.e. $h \coloneq h(x, o)$. In that case, the CBF constraint \cref{eq:cbf} can be written as:
\begin{equation} \label{eq:obs_depend_cbf_constr}
    \nabla_x h(x, o)^\top \dot{x} + \nabla_o h(x, o)^\top \dot{o} \geq - \alpha \left( h(x, o) \right).
\end{equation}
In this formulation, the main difficulty arises due to the unknown observation dynamics $\dot{o}$. This dynamics can be partially described by employing the sensor's model and predicting how the observation will change, e.g., due to the robot's motion \cite{dawson_learning_2022}. However, this approach does not account for the obstacles that are not yet observed because the information about the distribution of obstacles in the environment is not available a priori.

To mitigate unknown observation dynamics, we leverage the following two properties that appear often in practice. First, the observations are typically updated with a lower frequency compared to the state feedback, which means that a particular observation is constant over a considerable period of time. Second, when an observation is updated, the new information is usually introduced at the edges of the perception field\footnote{New information may also appear closer to the robot when occluded objects suddenly become visible, but this problem is beyond the scope of this paper.}. The only requirement is that the perception field is large enough so that the obstacles that appear on the observation boundary do not make the new state unsafe instantly. Based on that, we can interpret the CBF as being conditioned on the environment observation, i.e. $h \coloneq h(x|o)$, and the CBF constraint is defined as:
\begin{equation} \label{eq:obs_cond_cbf_constr}
    \dot{h}(x|o) = \nabla_x h(x|o)^\top \dot{x}\geq - \alpha \left( h(x|o) \right).
\end{equation}

The key practical implication of this interpretation is that the constraint formulation remains unchanged compared to the CBFs defined only for $x$. Additionally, by interpreting a CBF as a function conditioned on observation, we can directly use the existing numerical tools for HJ reachability analysis to compute the HJ value function for the given observation. Otherwise, it would be necessary to model and simulate observation dynamics, which would be impractical.

\begin{figure}[!tb]
    \centering
    \includegraphics[width=1\linewidth]{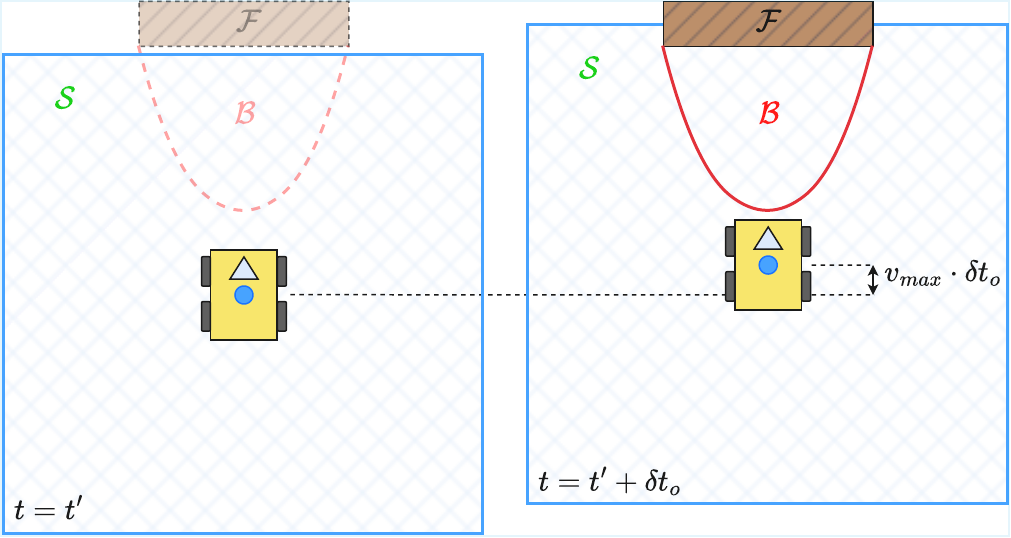}
    \caption{Illustration of the critical moment when the new observation occurs. The robot must be outside of the BRT $\mathcal{B}$ corresponding to the suddenly observed failure set $\mathcal{F}$. This can be achieved by considering all aspects such as the robot's maximal speed, observation size and update rate, maximal expected size of an obstacle, and the shape of the corresponding BRT.}
    \label{fig:new_obs_illustr}
\end{figure}

In essence, a new CBF is generated when a new observation becomes available, which is the most critical moment. Whether the system stays safe when the new observation occurs depends on multiple factors such as the observation update rate, the maximal speed of the robot, the size of the perception field, the maximal size of an obstacle and the corresponding BRT shape. As mentioned earlier, it is required that the robot stay within the safe set, i.e., outside of the BRT, when the new observation occurs and the CBF is updated. We illustrate this in \cref{fig:new_obs_illustr} where the robot moves with the maximal speed $v_{max}$ and the observation is updated every $\delta t_o$. In the worst case, the robot will move maximally $v_{max} \delta t_o$ between the two observations and the obstacle will penetrate the perception field (blue region) for the same length in the new observation at $t=t'+\delta t_o$. It is then important that the robot is outside of the obstacle's BRT $\mathcal{B}$ at that moment in order to continue safe motion.  

\subsection{Observation-conditioned residual neural CBF}
\label{subsec:on_cbf}

Among the most important properties of a CBF is the size of the safe set $\mathcal{S}$ associated with that CBF. This is our main motivation for using the HJ value function as a CBF since this function recovers the optimal safe set for the given system \cite{choi_robust_2021, tonkens_refining_2022, so_how_2024}. However, computing the HJ value function in real time is generally intractable and therefore we use a learning-based approach, which allows us to approximate the value function based on available observations and then employ that approximation as an observation-conditioned CBF, i.e., $\hat{h}(x|o)\coloneq \hat{V}(x|o)$.

First, we recall that the HJ value function obtained by solving HJB-VI \cref{eq:hjb_vi} is less than or equal to the corresponding failure function \cite{derajic_residual_2025}. In our context, since the failure function is equal to the SDF, we have that $h(x|o) \leq d(x|o), \, \forall x$, which means that the CBF can be obtained by subtracting a nonnegative residual function from the SDF:
\begin{equation}
    h(x|o) = d(x|o) - r(x|o) \,\ s.t. \,\ r(x|o) \geq 0, \forall x.
\end{equation}
Also, we note that a continuous SDF approximation $\hat{d}(x|o)$ can be derived by a simple interpolation of a discretized SDF $\bar{d}$, which in turn can be efficiently computed from an observation such as an occupancy grid. As a result, the formulation above allows us to learn only the residual component instead of the complete CBF.

We approximate the residual using a neural network with parameters $\Theta$ and denote it with $\hat{r}_{\Theta}(x | o)$. To ensure a nonnegative value for all inputs, we apply a nonnegative activation at the model's output. A straightforward option would be the ReLU function defined as $\max(0, z)$. However, the gradient of the function is always equal to 0 for negative inputs, which can negatively impact the training process. Also, the activation should be continuously differentiable, so we opt for the softplus function:
\begin{equation} \label{eq:sofplus}
    \eta(z) = \log \left( 1 + \exp(z) \right).
\end{equation}
In addition, given that the approximation of the CBF gradient $\nabla \hat{h}(x | o)$, and therefore $\nabla \hat{r}_{\Theta}(x | o)$, is required for the CBF constraint \cref{eq:obs_cond_cbf_constr}, we use the multilayer perceptron (MLP) model with sinusoidal activation functions. Those models have been proven to work well for applications where the gradient of the approximated function has to be accurately approximated as well \cite{sitzmann_implicit_2020, bansal_deepreach_2021}.

However, instead of directly learning parameters $\Theta$, we take the hypernetwork-based architecture which involves two neural networks - the hypernetwork and the main network. The hypernetwork is a large and expressive trainable model that parametrizes the main network for the given hypernetwork input. On the other hand, the main network is a simple model that approximates the target function for that particular input of the hypernetwork. In our context, the input to the hypernetwork is the discretized SDF $\bar{d}$ obtained from the observation $o$, while the main network is the previously described MLP model approximating the residual function for the given observation. This architecture is particularly suitable for our observation-conditioned interpretation of CBF safety filtering because the hypernetwork essentially conditions the main network to approximate the residual for the given observation. Also, this architecture has been demonstrated to be more efficient than a single-network model because the hypernetwork is inferred only once when the new observation becomes available, which happens at a lower rate, while the main network is queried for its value and gradient with a much higher frequency \cite{derajic_learning_2025}. Moreover, the two models can be designed almost completely separately; the only requirement is that the number of outputs of the hypernetwork is equal to the number of parameters of the main network. 

\begin{figure*}[!tb]
    \centering
    \includegraphics[width=\textwidth]{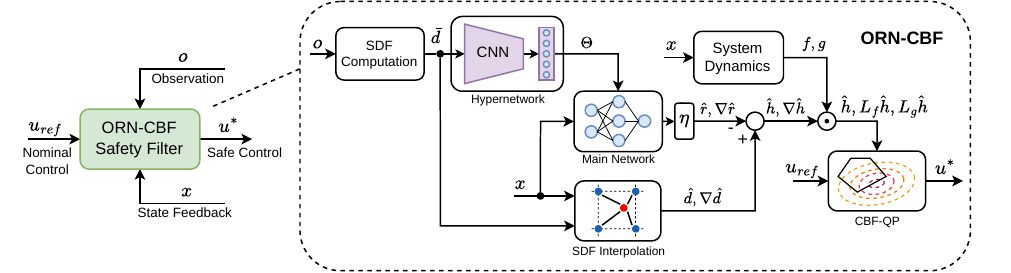}
    \caption{A detailed architecture of the ORN-CBF safety filter. Based on the observation $o$, a discretized SDF $\bar{d}$ is computed, which is then fed into the hypernetwork and the SDF interpolation function. The hypernetwork parametrizes the main network, which approximates the nonnegative residual $\hat{r}$ and its gradient $\nabla \hat{r}$. The approximate CBF value $\hat{h}$ and gradient $\nabla \hat{h}$ are obtained by subtracting $\hat{r}$ and $\nabla \hat{r}$ from interpolated SDF value $\hat{d}$ and gradient $\nabla \hat{d}$, respectively. In the end, $L_f\hat{h}$ and $L_g\hat{h}$ are computed based on $\nabla \hat{h}$ and the system dynamics and together with $\hat{h}$ provided to the CBF-QP safety filter.}
    \label{fig:orn_cbf}
\end{figure*}

In summary, for the given observation $o$ (e.g. occupancy grid map), we first compute a discretized SDF $\bar{d}$ and provide it to the hypernetwork and to the interpolation function. The hypernetwork outputs parameters $\Theta$ that parameterize the main network representing residual $\hat{r}_{\Theta}(x | o)$, the interpolation function represents $\hat{d}(x | o)$, while the approximated CBF is obtained as
\begin{equation} \label{eq:approx_obs_cond_cbf}
    \hat{h}(x|o) = \hat{d}(x|o) - \hat{r}_{\Theta}(x | o).
\end{equation}
By implementing the main network and interpolation function in a framework that supports automatic differentiation (e.g. PyTorch or CasADi), the gradients $\nabla \hat{d}(x|o)$ and $\nabla \hat{r}_{\Theta}(x | o)$ can be efficiently obtained for the given state $x$. The CBF gradient $\nabla \hat{h}(x|o)$ is then used to compute the Lie derivatives $L_f\hat{h}(x|o)$ and $L_g\hat{h}(x|o)$, which are finally provided to the CBF-QP together with $\hat{h}(x|o)$. We call this method observation-conditioned residual neural CBF (ORN-CBF) and the complete architecture is visualized in \cref{fig:orn_cbf}.

To train the hypernetwork for the proposed ORN-CBF method, we resort to the supervised approach, which requires both inputs and the corresponding target outputs. In our setting, we have two sets of inputs: a set of discretized SDFs (inputs to the hypernetwork) and a set of state space grid points (inputs to the main network). The target outputs are HJ value functions numerically computed over the same state space grid, representing the target CBFs paired with the corresponding observations. As a loss function, we use the radially weighted MSE (RWMSE) loss function proposed in \cite{derajic_learning_2025}, which improves approximation accuracy near the zero-level sets of the value function. 

Taking into account that the robot travels a relatively short distance in space between two observations, instead of learning the HJ value function over the complete observable space like in \cite{derajic_learning_2025}, we can learn the HJ value function only for the surrounding positions reachable from the current position before the observation is updated. For example, the environment observation in our experiments is a square-shaped occupancy grid centered at the robot's position at the time when the observation occurs. Assuming that the robot's maximal speed is $v_{max}$ in all directions and that the observations are updated every $\delta t_o$, the robot can reach only positions within the circle of radius $v_{max}\delta t_o$ before the observation is recentered at the robot's new position. Therefore, it is enough to approximate the HJ value function only for the region that includes those reachable positions\footnote{This holds only for the positional coordinates of the state vector, while the rest of the state space should be covered completely.}. The key practical benefit of this insight is a more efficient learning process since the neural CBF is trained to approximate the HJ value function only for this patch, and not the complete observed space. This results in lower memory requirements, faster training and less complex models.

\section{EXPERIMENTAL RESULTS}
\label{subsec:experiments}

In this section, we first introduce the considered robot models, followed by details on the hypernetwork training process and an analysis of the results obtained from simulation and hardware experiments\footnote{Code will be released after the paper is published.}.

\subsection{Robot Models}
\label{subsec:robot_models}

 The first type of robot is a ground robot modeled as a Dubins car, which is a first-order unicycle model with constant linear speed and limited angular speed. In practice, this model describes systems for which linear speed is not controllable or its change over time is negligible. Some examples include vehicles with faulty braking systems, forklifts transporting unstable objects, fixed-wing aircraft, marine vessels, etc. The state vector ${x = [p_x, \, p_y, \, \theta]^\top}$ includes positional coordinates $(p_x, p_y)$ and orientation angle $\theta$, while the control input is only the angular speed, i.e. ${u = \omega}$. The dynamics equation is
\begin{equation} \label{eq:dubins_car}
    \dot{x} = \left[
        v \cos(\theta), \,
        v \sin(\theta), \,
        \omega
    \right]^\top,
\end{equation}
where $v = 1.0$ m/s and ${\omega \in [-0.5, 0.5]}$ rad/s.

The second type of robot is a quadcopter modeled as a 2D double integrator with limited speed and acceleration. Such models are used in practice for quadcopters operating at constant heights with relatively low acceleration capabilities. Some examples are quadcopters transporting heavy objects at construction sites, in agriculture, or for delivery. The state of the robot is ${x = [p_x, \, p_y, \, v_x, \, v_y]^\top}$, where $(p_x, p_y)$ are positional coordinates and $v_x, v_y$ are velocities. The control vector $u = [a_x, \, a_y]^\top$ consists of linear accelerations, while the dynamics equation is
\begin{equation} \label{eq:2d_doub_int}
    \dot{x} = \left[
        v_x, \,
        v_y, \,
        a_x, \,
        a_y
    \right]^\top,
\end{equation}
where $v_x, v_y \in [-2, 2]$ m/s and $a_x, a_y \in [-1, 1]$ m/s$^2$.

\subsection{Model Training}
\label{subsec:model_training}

As described in \cref{sec:methodology}, we train the model in a supervised fashion, and the first part is to obtain a suitable dataset consisting of inputs and target outputs. For the hypernetwork inputs, we first collect a set of observations in the form of occupancy grid maps and compute the corresponding discretized SDFs using the method described in \cite{maurer_linear_2003}. For the ground robot, we collect 2,500 costmaps of size 6$\times$6 m with resolution 0.06 m from a 3D model of a warehouse environment. Similarly, we collect 2,000 costmaps of size 6$\times$6 m with resolution 0.075 m from a forest-like 3D environment for the quadcopter model. We additionally augment both datasets by flipping horizontally and rotating for 90$^{\circ}$, 180$^{\circ}$ and 270$^{\circ}$ each original and flipped costmap, resulting in 20,000 costmaps for the ground robot and 16,000 costmaps for the quadcopter.

Next, we define state space grids over which the HJ value functions are computed and which are used as the input to the main network. We define state space grids of shape ${100\times100\times30}$ for the Dubins car and ${80\times80\times20\times20}$ for the 2D double integrator model. To compute the HJ value functions, we use the \textit{hj\_reachability}\footnote{\url{https://github.com/StanfordASL/hj_reachability}} Python package. The computation takes $\sim$1.17 h in the first case, and $\sim$16.3 h in the second case, on a single RTX3090 GPU. Then, we store the value functions and grid points only for the subset of the initial grid according to the explanation in the last paragraph of \cref{sec:methodology}. Concretely, for both models, we store only the central patch in the positional subspace of size ${16\times16}$, meaning that we store ${16\times16\times30}$ grid points and the corresponding function values for the Dubins car and ${16\times16\times20\times20}$ for the 2D double integrator.

\begin{figure}[!tb]
    \centering
    \includegraphics[width=1\linewidth]{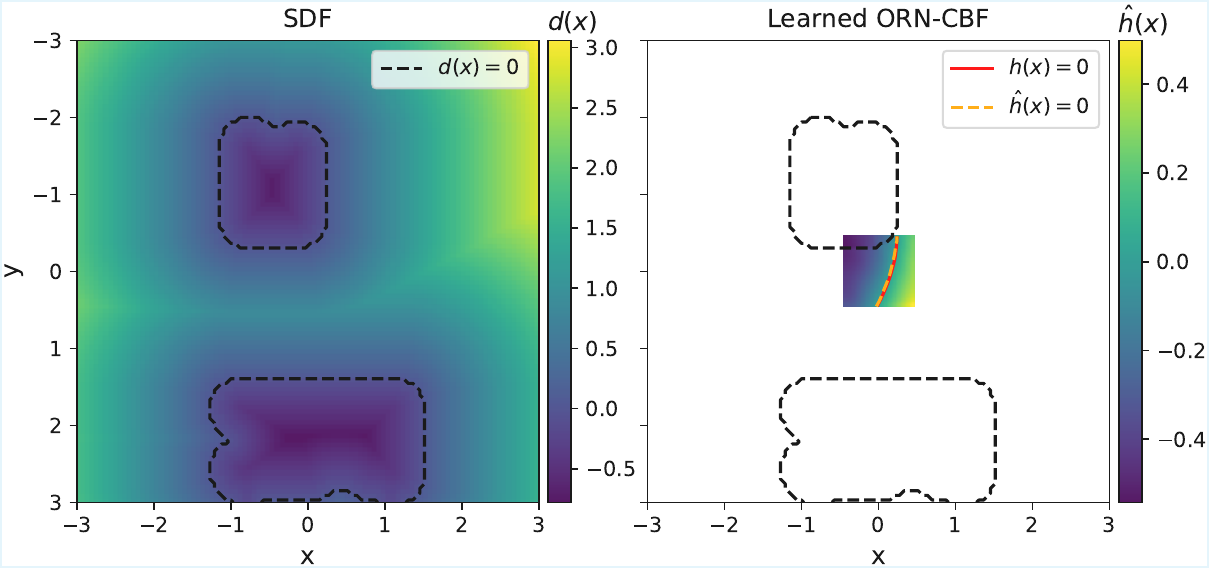}
    \caption{A slice of a learned ORN-CBF for the Dubins car for $\theta = \frac{\pi}{2}$ rad.}
    \label{fig:sample_cbf_dubins_car}
\end{figure}

After aggregating the datasets, we proceed with the model training. For both types of robots, we use deep CNN models as the hypernetwork and MLP models as the main network. We use the same architecture of the main network in both cases: 32-32-32-16-16-16-8-8-8 units with sinusoidal activations in hidden layers and a single output with softplus activation, while the number of inputs is equal to the corresponding state dimension. During the training, at each epoch, a discretized SDF is propagated through the hypernetwork and the main network is parametrized with the predicted parameters. Then, the cropped state space grid is propagated through the main network to obtain the HJ value function predictions. Finally, the loss function is evaluated for the predicted and true value functions, and the parameters of the hypernetwork are updated via the gradient descent method. In this paper, we use the RWMSE loss function proposed in \cite{derajic_learning_2025} and train models for 100 epochs with a batch size of 16. We use the Adam optimizer with an initial learning rate of $10^{-4}$, which is decreased by a factor of 10 at epochs 85 and 95. The training process takes $\sim$0.5 h for the Dubins car and $\sim$2.9 h for the 2D double integrator on a single RTX3090 GPU, which is roughly 20 times faster compared to the method in \cite{derajic_learning_2025} under the same setting. In \cref{fig:sample_cbf_dubins_car} we visualize a sample input to the hypernetwork and the corresponding learned ORN-CBF for the Dubins car model.

\subsection{Simulation Experiments}
\label{subsec:sim_experiments}

\begin{figure}[!tb]
    \centering
    \includegraphics[width=\linewidth]{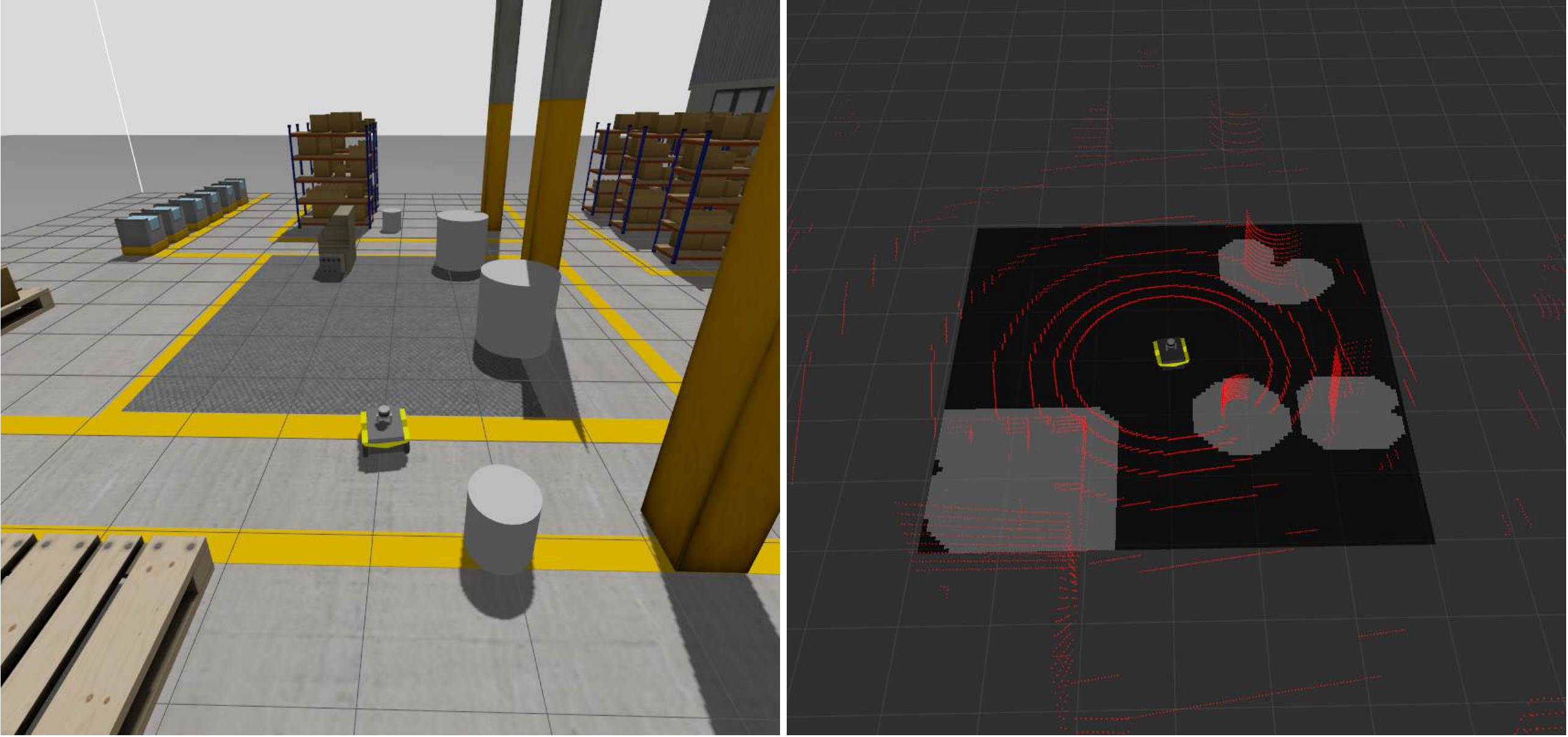}
    \caption{Left: Scene from the 3D simulation experiment with the ground robot in the warehouse environment. Right: The corresponding observation.}
    \label{fig:gr_sim_exp}
\end{figure}

The proposed ORN-CBF method is first evaluated in simulations implemented using the Gazebo 3D simulator. The ground robot, modeled as a Dubins car, is tested in a warehouse environment visualized in \cref{fig:gr_sim_exp}. As a nominal controller, we use a simple MPC local planner, which uses the SDF as a collision avoidance constraint (SDF-MPC). The ORN-CBF method is examined against SDF-MPC alone and more advanced MPC-based methods that have been proven to work well in practice for safe navigation in unknown environments: DCBF-MPC \cite{zeng_safety-critical_2021} and NTC-MPC \cite{derajic_learning_2025}. The related works on neural CBFs \cite{dawson_learning_2022} and \cite{harms_neural_2024} assume different observation modalities, and therefore are not directly applicable in our setting. Besides that, as an ablation study, we test a simpler variant of the proposed method, which directly approximates the HJ value function by the main network instead of the residual function. We call this method observation-conditioned neural CBF (ON-CBF). Even though this version has a less complex architecture, it does not provide any guarantees on the approximation error. As it is standard in practice, we use a linear function for $\alpha(\cdot)$, i.e. $\alpha(h) = kh, k \in \left[ 0, \infty \right)$.

\begin{figure}[!tb]
    \centering
    \includegraphics[width=0.75\linewidth]{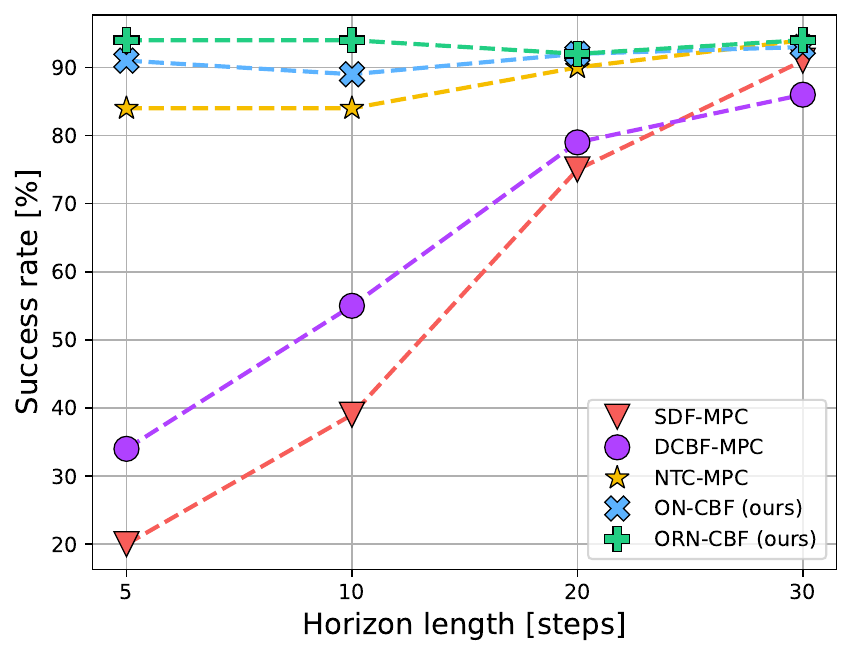}
    \caption{Obtained success rates for different lengths of the MPC prediction horizon for the Dubins car model. The results show that the proposed ORN-CBF method, and its simple version ON-CBF, outperform the MPC-based baselines for different horizon lengths.}
    \label{fig:gr_success_rate}
\end{figure}

All the MPC planners run with the control frequency of 20 Hz and use a sampling time of 0.1 s over the prediction horizon. The CBF-QP safety filters run at 200 Hz with $k=2.6$, and the costmap is updated at 20 Hz. The MPC planners and CBF-QP are implemented in CasADi \cite{andersson_casadi_2019} with the IPOPT NLP solver and qpOASES QP solver, respectively. We run 100 experiments where the task is to navigate the robot from the initial to the goal position based only on the local observation while avoiding collisions with randomly distributed obstacles. Since both performance and safety of the MPC planners largely depend on the prediction horizon, we repeat experiments for different horizon lengths. The obtained success rates are shown in \cref{fig:gr_success_rate}, demonstrating the advantage of using the proposed ORN-CBF safety filter. Moreover, even the simpler ON-CBF variant outperforms the baseline methods for short prediction horizon.

\begin{figure}[!tb]
    \centering
    \includegraphics[width=\linewidth]{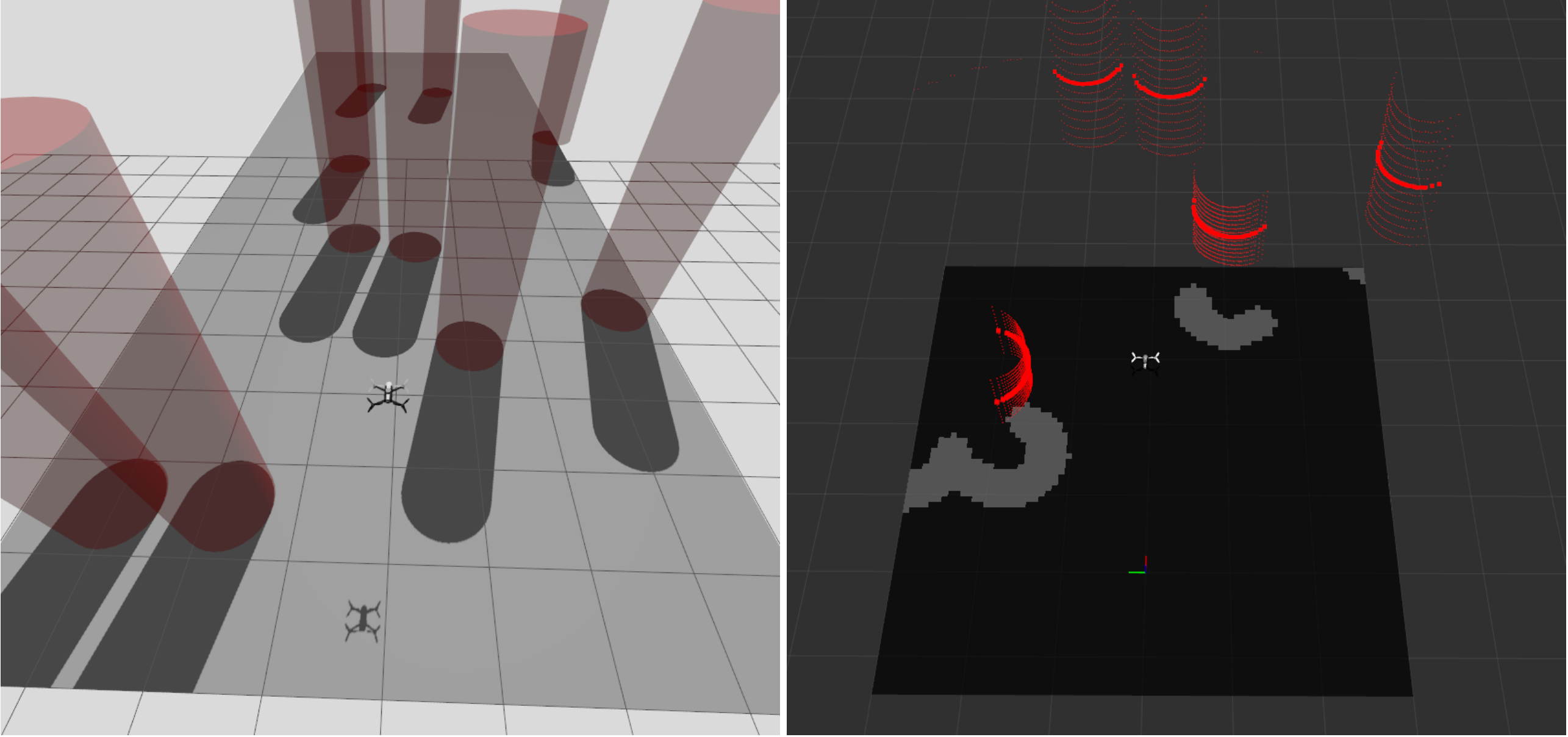}
    \caption{Left: Scene from the 3D simulation experiment with the quadcopter in the forest-like environment. Right: The corresponding observation.}
    \label{fig:qq_sim_exp}
\end{figure}

The quadcopter robot, modeled as a 2D double integrator, is evaluated in a forest-like environment shown in \cref{fig:qq_sim_exp}. As the nominal controller, we design a linear-quadratic regulator (LQR)  with $Q = \diag(10, 10, 1, 1)$ and $R = \diag(1, 1)$.

Instead of evaluating different nominal controllers, we examine the generalization properties of the learned CBFs. ORN-CBF and ON-CBF are trained using data from a forest-like environment with cylindrical obstacles of 0.5 m radius (in-domain environment). Both methods are then evaluated for the environment where obstacles' radii range from 0.2 m to 1.0 m (out-of-domain environment). The task is to navigate from the initial to the goal position, and in both cases, there are 200 scenarios with 10 obstacles randomly distributed in the area through which the robot navigates. The CBF-QP runs at 200 Hz and the costmap is updated at 5 Hz. 

\begin{figure}[!tb]
    \centering
    \includegraphics[width=1\linewidth]{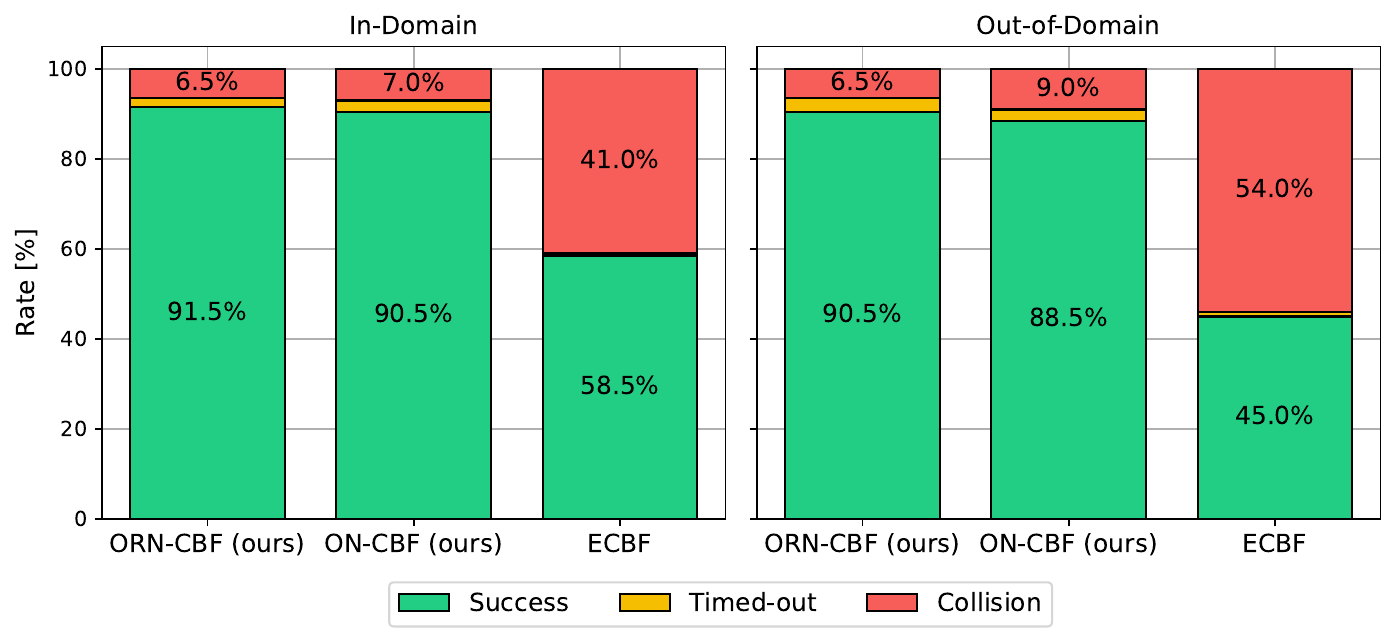}
    \caption{Results obtained from the simulation experiments with a quadcopter. The results indicate strong generalization properties of the proposed ORN-CBF and ON-CBF methods compared to a hand-tuned ECBF.}
    \label{fig:qq_sim_exp_results}
\end{figure}

\begin{figure}[!tb]
    \centering
    \includegraphics[width=1\linewidth]{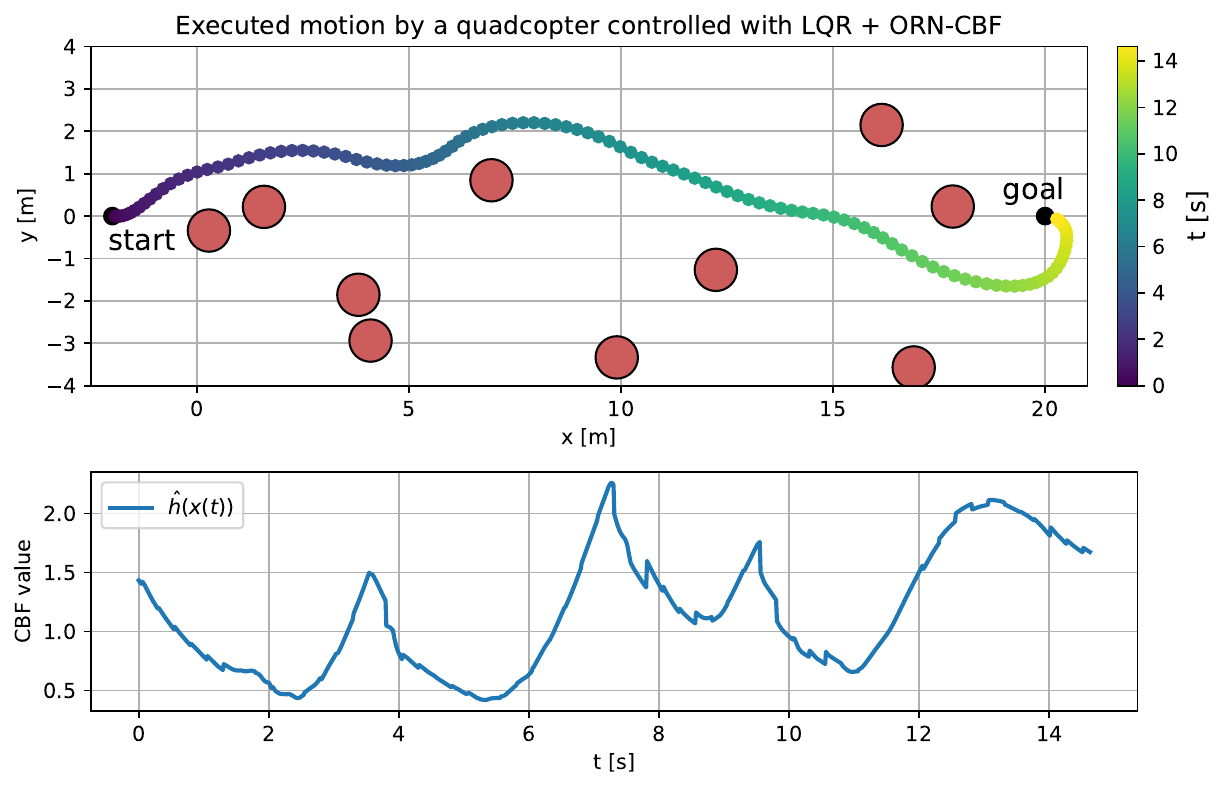}
    \caption{Up: Visualized motion of the quadcopter controlled by the LQR and the ORN-CBF safety filter. Down: The corresponding change of the ORN-CBF value over time.}
    \label{fig:qq_sample_traj}
\end{figure}

For comparison, we design an exponential CBF (ECBF) \cite{nguyen_exponential_2016} based on the SDF $d(x)$, which results in a second-order ECBF. The gain matrix ${K_{ECBF} = \left[1.0, \, 2.0 \right]}$ is tuned for the first environment and evaluated for both environments. The obtained results for $k=0.7$ are presented in \cref{fig:qq_sim_exp_results}, indicating solid robustness of the proposed CBF methods compared to a classical ECBF. Failing to achieve 100\% rate can be attributed to multiple factors such as learning error, signal delays, finite costmap size and resolution, etc. In \cref{fig:qq_sample_traj}, we visualize the motion of the quadcopter from a representative scenario in simulation experiments.

\subsection{Hardware Experiments}
\label{subsec:hardw_experiments}

\begin{figure}[!tb]
    \centering
    \includegraphics[width=1\linewidth]{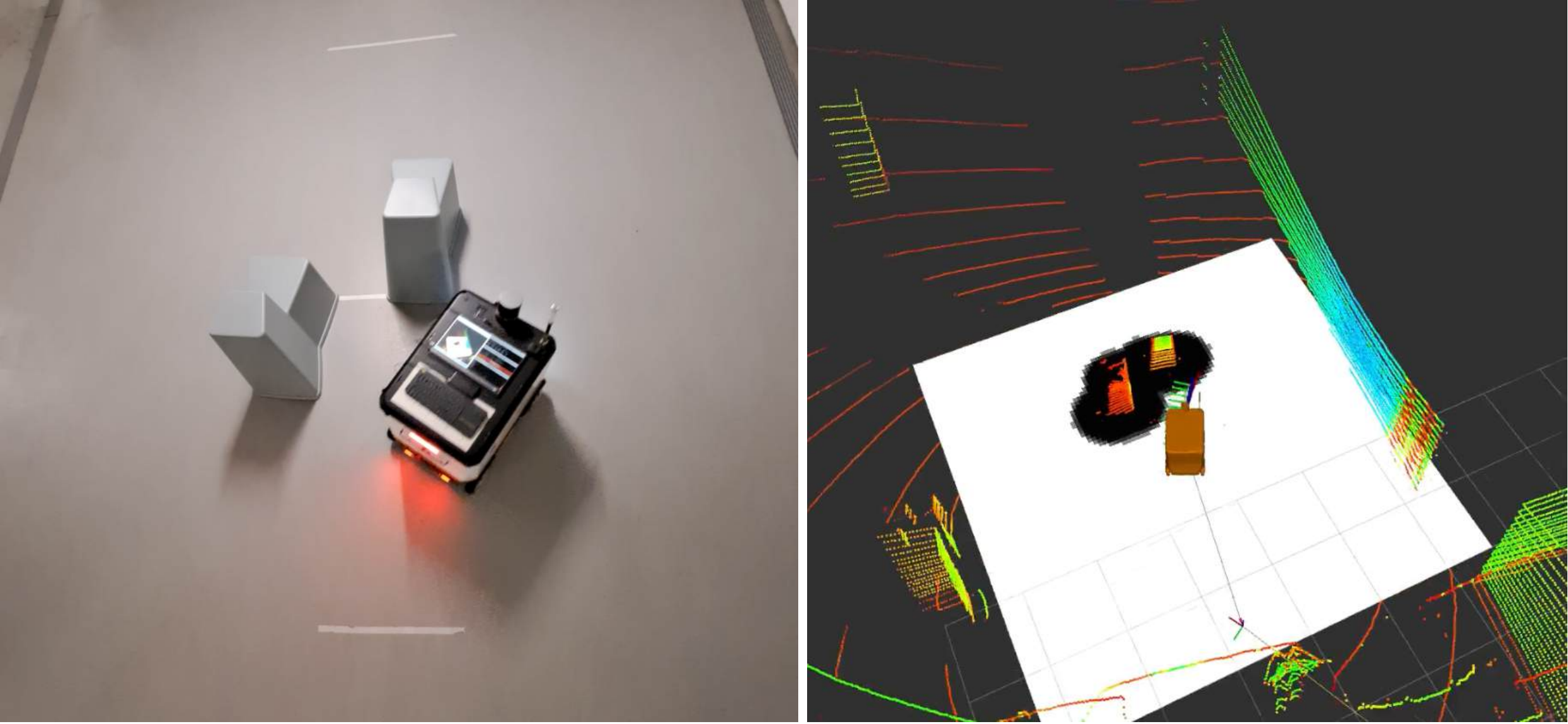}
    \caption{Left: Ground robot used in the hardware experiments. Right: Visualization of the local observation.}
    \label{fig:gr_hardw_exp}
\end{figure}

\begin{figure}[!tb]
    \centering
    \includegraphics[width=1\linewidth]{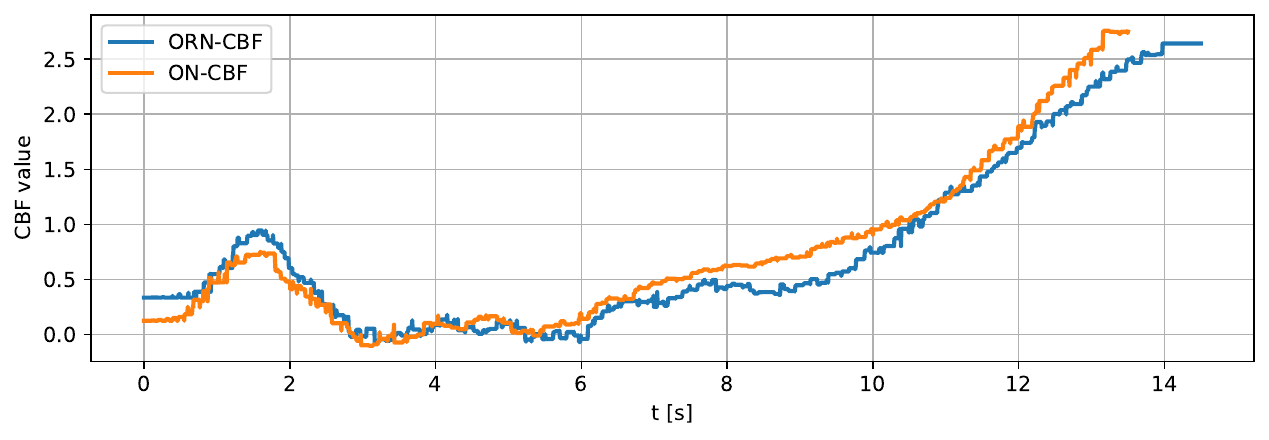}
    \caption{CBF values for the ORN-CBF and ON-CBF methods during one of the hardware experiments with the ground robot.}
    \label{fig:gr_cbf_value_hardw_exp}
\end{figure}

We perform hardware experiments with a ground robot modeled as a Dubins car, shown in \cref{fig:gr_hardw_exp}. To evaluate sim-to-real robustness, we use the same ORN-CBF and ON-CBF models trained on synthetic data. We compare against the same MPC-based baselines using a prediction horizon of 10 steps, a sampling time of 0.1 s, and a control frequency of 20 Hz. The CBF safety filters run at 200 Hz, and all computations are performed onboard. For each method, we run 10 experiments with random obstacle distributions. The results in \cref{tab:gr_hardw_exp_success_rates} show that the proposed safety filters perform significantly better than the baselines. \cref{fig:gr_cbf_value_hardw_exp} illustrates the CBF value over time for ORN-CBF and ON-CBF in one representative scenario. The CBF values occasionally dip slightly below zero due to model mismatch and measurement noise, yet collisions are avoided thanks to a small buffer zone that accounts for these uncertainties.

\begin{table}[!tb]
    \centering
    \caption{Results from the Hardware Experiments with the Dubins Car}
    \label{tab:gr_hardw_exp_success_rates}
    \renewcommand\tabularxcolumn[1]{m{#1}}
    \begin{tabularx}{\linewidth}{*{6}{>{\centering\arraybackslash}X}}
         & SDF-MPC & DCBF-MPC & NTC-MPC & ORN-CBF & ON-CBF \\
        \hline
        \noalign{\vskip 1mm} 
        \makebox[0pt][c]{Success rate} & 20\% & 40\% & 70\% & \textbf{100\%} & \textbf{100\%} \\
    \end{tabularx}
\end{table}

We also conduct hardware experiments with a Crazyflie quadcopter equipped with 4 laser-based distance sensors. The local costmap is created by a constant rotation of the quadcopter around the z-axis during motion. We deploy the same LQR nominal controller and learned ORN-CBF safety filter used in simulations, while the costmap is updated at the rate of 10 Hz and computation is performed offboard. The experiments illustrate successful obstacle avoidance when using the proposed ORN-CBF method, despite the challenging hardware setup (see the supplemental video).

\section{DISCUSSION ON LIMITATIONS}
\label{sec:limitations}

We use numerical tools for HJ reachability during data generation, but these tools are typically not applicable to systems with more than six state dimensions. To improve scalability, one can use the self-supervised approach for model training as in \cite{bansal_deepreach_2021}, or compute approximate HJ value functions using the MPC framework \cite{feng_bridging_2025}.

On the other hand, the assumption about a static environment might limit the scope of use for our method. A solution would be to employ HJ reachability analysis for time-varying failure sets as in \cite{derajic_residual_2025}. This would require a different data generation process and a hypernetwork that can process temporal data instead of only the current observation. 

\section{CONCLUSIONS}

We propose a novel learning-based method, called ORN-CBF, for designing observation-conditioned CBFs in unknown environments. The resulting CBF represents an HJ value function and approximately recovers the maximal safe set for the current observation. By learning only the residual component of the HJ value function, we guarantee that the predicted safe set never includes the observed failure set. Moreover, the proposed hypernetwork-based architecture enables a computationally efficient safety filter.

We extensively evaluate ORN-CBF in simulations on a ground robot and a quadcopter. The results show improved success rates and robust out-of-domain performance compared to the baselines. Hardware experiments further demonstrate that the safety filters can be readily deployed on different real robots. Future work will extend the method to dynamic environments and higher-dimensional systems.


\bibliographystyle{IEEEtran}
\bibliography{references}

\end{document}